\documentclass[letterpaper, 10 pt, conference]{ieeeconf}

\IEEEoverridecommandlockouts                              

\usepackage{cite}
\usepackage{listings}
\usepackage{float}
\usepackage[dvipdfmx]{graphicx}
\usepackage{amssymb}
\usepackage{latexsym}
\usepackage{amsfonts}
\usepackage{url}
\usepackage{comment}
\usepackage{algorithm}
\usepackage{algpseudocode}

\newcommand{\FIG}[3]{
\begin{minipage}[b]{#1cm}
\begin{center}
\includegraphics[width=#1cm]{#2}\\
{\scriptsize #3}
\end{center}
\end{minipage}
}

\newcommand{\FIGU}[3]{
\begin{minipage}[b]{#1cm}
\begin{center}
\includegraphics[width=#1cm,angle=180]{#2}\\
{\scriptsize #3}
\end{center}
\end{minipage}
}

\newcommand{\FIGm}[3]{
\begin{minipage}[b]{#1cm}
\begin{center}
\includegraphics[width=#1cm]{#2}\\
{\scriptsize #3}
\end{center}
\end{minipage}
}

\newcommand{\FIGR}[3]{
\begin{minipage}[b]{#1cm}
\begin{center}
\includegraphics[angle=-90,width=#1cm]{#2}
\\
{\scriptsize #3}
\vspace*{1mm}
\end{center}
\end{minipage}
}

\newcommand{\FIGRpng}[5]{
\begin{minipage}[b]{#1cm}
\begin{center}
\includegraphics[bb=0 0 #4 #5, angle=-90,clip,width=#1cm]{#2}\vspace*{1mm}
\\
{\scriptsize #3}
\vspace*{1mm}
\end{center}
\end{minipage}
}

\newcommand{\FIGpng}[5]{
\begin{minipage}[b]{#1cm}
\begin{center}
\includegraphics[bb=0 0 #4 #5, clip, width=#1cm]{#2}\vspace*{-1mm}\\
{\scriptsize #3}
\vspace*{1mm}
\end{center}
\end{minipage}
}

\newcommand{\FIGtpng}[5]{
\begin{minipage}[t]{#1cm}
\begin{center}
\includegraphics[bb=0 0 #4 #5, clip,width=#1cm]{#2}\vspace*{1mm}
\\
{\scriptsize #3}
\vspace*{1mm}
\end{center}
\end{minipage}
}

\newcommand{\FIGRt}[3]{
\begin{minipage}[t]{#1cm}
\begin{center}
\includegraphics[angle=-90,clip,width=#1cm]{#2}\vspace*{1mm}
\\
{\scriptsize #3}
\vspace*{1mm}
\end{center}
\end{minipage}
}

\newcommand{\FIGRm}[3]{
\begin{minipage}[b]{#1cm}
\begin{center}
\includegraphics[angle=-90,clip,width=#1cm]{#2}\vspace*{0mm}
\\
{\scriptsize #3}
\vspace*{1mm}
\end{center}
\end{minipage}
}

\newcommand{\FIGC}[5]{
\begin{minipage}[b]{#1cm}
\begin{center}
\includegraphics[width=#2cm,height=#3cm]{#4}~$\Longrightarrow$\vspace*{0mm}
\\
{\scriptsize #5}
\vspace*{8mm}
\end{center}
\end{minipage}
}

\newcommand{\FIGf}[3]{
\begin{minipage}[b]{#1cm}
\begin{center}
\fbox{\includegraphics[width=#1cm]{#2}}\vspace*{0.5mm}\\
{\scriptsize #3}
\end{center}
\end{minipage}
}

\newcommand{\editage}[2]{#1}

\begin{document}

\title{\LARGE \bf%
DRIP: Discriminative Rotation-Invariant Pole Landmark Descriptor for 3D LiDAR Localization
}

\author{Dingrui Li, Dedi Guo, and Kanji Tanaka
\thanks{D. Li, D. Guo, and K. Tanaka are with Department of Engineering, University of Fukui, Japan. {\tt\small tnkknj@u-fukui.ac.jp}}
}

\maketitle

\begin{abstract}
In 3D LiDAR-based robot self-localization, pole-like landmarks are gaining popularity as lightweight and discriminative landmarks. This work introduces a novel approach called ``discriminative rotation-invariant poles," which enhances the discriminability of pole-like landmarks while maintaining their lightweight nature. Unlike conventional methods that model a pole landmark as a 3D line segment perpendicular to the ground, we propose a simple yet powerful approach that includes not only the line segment's main body but also its surrounding local region of interest (ROI) as part of the pole landmark. Specifically, we describe the appearance, geometry, and semantic features within this ROI to improve the discriminability of the pole landmark. Since such pole landmarks are no longer rotation-invariant, we introduce a  novel rotation-invariant convolutional neural network that automatically and efficiently extracts rotation-invariant features from input point clouds for recognition. Furthermore, we train a pole dictionary through unsupervised learning and use it to compress poles into compact pole words, thereby significantly reducing real-time costs while maintaining optimal self-localization performance. Monte Carlo localization experiments using publicly available NCLT dataset demonstrate that the proposed method improves a state-of-the-art pole-based localization framework.
\end{abstract}

\section{Introduction}

Visual self-localization aims for autonomous mobile robots to estimate their position by matching measurement data from visual sensors to a pre-built map. Historically, onboard cameras have been widely used as visual sensors, but recently, there has been growing interest in applying onboard LiDAR. Previous studies have focused on extracting discriminative features (point-based, voxel-based, pillar-based) from 3D point clouds suitable for memory and matching. These studies have shown that 3D LiDAR point cloud data is robust to changes in lighting conditions and can achieve more accurate positioning. However, those 3D LiDAR point cloud features are typically large in volume, posing significant spatial and temporal costs for storage and matching. To address these challenges, simple landmark-based representations have recently gained attention. Notably, the seminal work by Dong et al. \cite{dong2023online} has demonstrated the feasibility of lightweight localization using pole-like landmarks. However, landmark-based methods discard point cloud data not belonging to the landmarks, potentially resulting in representations significantly less discriminative compared to the original point cloud.

This work aims to achieve fast and accurate localization based on pole landmarks by enhancing their discriminability while maintaining their lightweight nature. Such localization can significantly increase the scalability of landmark-based localization, making it applicable to large cities and planetary-scale applications. However, efficient localization based on pole-like landmarks is a non-trivial problem. First, the spatial distribution of pole-like landmarks exhibits significant bias. For example, in forest environments, trees densely populate the space as pole-like landmarks, whereas, in suburban environments, such landmarks are rare. Clearly, adaptive sampling to the spatial distribution of landmarks is crucial. Secondly, not all landmarks observed during map construction are necessarily observed during localization, as the viewpoints of the robot building the map and the robot using the map are not always identical. Moreover, landmarks may appear or disappear between the time the map is built and the time it is used. Third, and most importantly, while pole objects are easy to detect, they themselves have poor appearance features and are difficult to distinguish from one another.

To balance discriminability and lightweight nature is a non-trivial issue, and to address this, we make three contributions. (1) Discriminability: To improve the discriminability of a pole-like landmark, we introduce a discriminative pole descriptor that describes the region-of-interest (ROI) around the pole, rather than simply modeling the pole as a vertical line segment perpendicular to the ground. (2) Rotation-invariance: Since such pole landmarks are no longer rotation-invariant, we introduce a novel rotation-invariant convolutional neural network that automatically and efficiently extracts rotation-invariant features from input point clouds for recognition. (3) Lightweight: Although the feature vectors from the deep pole classifier are discriminative and rotation-invariant, they remain high-dimensional and unsuitable for real-time recognition. Therefore, we use a pole dictionary pre-trained through unsupervised learning to vector-quantize a high-dimensional pole landmark descriptor into a pole-word. Our experiments demonstrate that the proposed method improves state-of-the-art pole-based localization using the Monte Carlo localization algorithm in a challenging cross-domain setup.

\section{
Related Work
}

{\bf Pole landmark-based localization.} With advancements in wide-area, high-precision LiDAR sensors and 3D point cloud recognition technologies based on deep learning, the recognition performance of pole objects has dramatically improved. By detecting lightweight and discriminative pole objects with low false positives/negatives, it has been demonstrated that efficient and accurate localization can be achieved. Our work is most related to the pole landmark-based localization method proposed by Dong et al. \cite{dong2023online}, which will serve as a baseline for performance comparison in our experiments. However, those existing approaches typically model a pole landmark as a 3D line segment perpendicular to the ground, thereby largely ignoring the discriminative information inherent in the point cloud. In contrast, our study models a pole landmark by incorporating not only its main body but also the rich contextual information within its local region-of-interest (ROI), aiming to significantly enhance its discriminative capability while maintaining the lightweight nature of the localization framework

{\bf Classification of pole objects.} Looking beyond the field of localization, ambitious research efforts are underway aimed not only at detecting poles but also at recognizing them. Notably, the seminal work in \cite{9564759} addresses the problem of classifying pole objects encountered by LiDAR-equipped vehicles into several known semantic classes (e.g., trees, streetlight) in autonomous driving scenarios. They formulate the training of their classifier as supervised learning using a pre-prepared annotated training set of pole objects. Supervised learning generally promises higher predictive accuracy compared to unsupervised or self-supervised learning, so this approach can be directly applicable to our pole landmark recognition. However, a well-known limitation of supervised learning is its inability to handle domain gaps between the dictionary and target domains. In open-world scenarios targeted by autonomous mobile robots, the distribution of poles is highly environment-dependent, and some pole classes encountered in the dictionary domain may rarely appear in the target domains. In contrast, our work allows the robot to automatically define the set of pole classes on the fly using an unsupervised approach based on pole objects encountered in the training environment, making it adaptable to previously unknown pole classes and class distributions.

{\bf Sparse landmark maps.} The landmark maps used in this work consist of spatially sparse pole landmarks, categorizing them as sparse landmark maps. For instance, the work in \cite{7785100} showcases robustness and scalability in processing city-scale point cloud data, effectively categorizing landmark buildings through spatial analysis and machine learning techniques based on their saliency characteristics. However, if the set of landmarks differs between the training and test environments, such saliency-based predictions may encounter issues due to domain gap when the model predicts saliency. In contrast, our work deals with a more challenging situation where a sparse set of landmarks is randomly selected. Thus, our approach is orthogonal and complementary to saliency-based approaches. Another related line of research is map matching, which aligns landmark configurations between the map and observations. For example, the work in \cite{neira2003linear} uses a RANSAC-based method to search for correspondences of landmark sets under 2D rotation and translation, achieving success. However, this RANSAC step incurs significant computational costs proportional to the environment's scale. In contrast, our work can omit the RANSAC step, resulting in a scalable method relative to the environment's scale.

\section{%
Method
}

\subsection{
Pole Extraction and Preprocessing
}

In this study, we utilized the NCLT dataset, which encompasses a wide range of indoor and outdoor sensor data. During the preprocessing phase, we implemented a pole extraction algorithm based on range images, as in \cite{dong2021online}. This method employs clustering analysis and geometric constraints to identify pole-like objects.

Initially, the algorithm clusters pixels in the range images into various small regions based on their range values. Subsequently, it applies a set of heuristic rules, such as aspect ratio and minimum height thresholds, to filter potential pole candidates. Finally, each candidate cluster is analyzed to determine the pole's center and radius by fitting the two-dimensional coordinates of the cluster. This process effectively identifies poles that stand out from the background and conform to specific geometric characteristics, facilitating accurate localization and mapping of pole-like structures within the dataset.

Following the segmentation, the pole data undergo normalization to ensure consistency in the input data.
\begin{itemize}
\item
Centroid calculation: The centroid  $c=n^{-1}\sum_{i=1}^np_i$ of a point cloud scan is computed where $p_i$ represents the $i$-th point in the point cloud, and $n$ denotes the total number of points within the cloud.
\item
Centroid alignment: Each point's coordinates are adjusted by subtracting the centroid coordinates, ensuring that the transformed point cloud's centroid is at the origin $p_i^{(1)}=p_i-c$.
\item
Normalization: The maximum distance from any point in the cloud to the origin is calculated. Subsequently, the entire point cloud is scaled by this maximum distance to normalize its size within a unit sphere:
$p^{(2)}_i=p^{(1)}_i(\max(\sum_i || p^{(1)} ||^2))^{-1/2}$.
\end{itemize}

\subsection{%
Model Training and Feature Extraction
}

To effectively handle unlabeled point cloud data, this study employs an enhanced RIConv network \cite{zhang2022riconv++}, which features rotational invariance, making it suitable for processing point clouds obtained from various angles. In an unsupervised learning setting, the Chamfer Distance (CD) is utilized as the loss function to effectively guide the model in learning feature representations that can best reconstruct the input point clouds. The CD loss encourages the model's output to geometrically resemble the actual input as closely as possible:
\begin{equation}
L^{CD} = \frac{1}{|X|} \sum_{x\in X} \min_{y\in Y} || x-y ||^2 + \frac{1}{|Y|} \sum_{y\in Y} \min_{x\in X} || y-x ||^2
\end{equation}
In this model, $X$ and $Y$ represent the predicted and reference point clouds, respectively. The Chamfer Distance (CD) loss function compels the network to generate point clouds whose structures closely match those of the target point clouds, thereby facilitating high-quality point cloud reconstruction. Global features are extracted from the signal flow before the fully connected layers and utilized as the foundation for subsequent k-means clustering analysis.

The model was trained for 400 epochs. After training, it was used for feature extraction. This approach allows us to effectively leverage unlabeled data, enhancing the model's ability to extract robust features.

\subsection{%
K-means Clustering and Pole Dictionary
}

The extracted features are fed into the k-means clustering algorithm, which optimizes the clustering centers by minimizing the intra-cluster variance. The mathematical formulation for this is:
\begin{equation}
\arg \min_S \sum_{i=1}^k \sum_{x\in S_i} || x-\mu_i ||^2
\end{equation}
where $S$ represents the set of clusters divided into $k$ groups, and $\mu_i$ is the center of cluster $S_i$. 
Through this process, the algorithm generates pseudo-labels that reflect the characteristics of the poles, providing essential support for the subsequent localization algorithms. This approach enables a structured utilization of unlabeled data by inferring meaningful labels that can be used to enhance the accuracy and robustness of the localization process.
We refer to these pseudo-labels as ``pole words." Our method uses a k-means classifier as a ``pole dictionary," which is employed to map a pole's ROI descriptor to a pole word.

\subsection{%
Improvements in Localization Algorithm
}

In this study, we have employed the Monte Carlo localization (MCL) framework in \cite{dong2023online} and have made significant improvements to adapt to challenging environmental conditions, particularly in scenarios with dense pole-like objects. The main enhancements include:

(1) Use of pole-words: Features are extracted from point cloud data through clustering analysis and used to generate a pole-word, which compactly represents the feature of a pole-like object. During the localization process, these pole-words serve as a crucial factor to enhance the algorithm's environmental perception.

(2) Pole-word consistency check: We have introduced a pole-word consistency check to improve the self-localization accuracy. This method involves comparing the environmental features near the predicted position with the currently observed pole-words, effectively distinguishing similar pole-like objects when viewed from varying angles. The mathematical expression for the pole-word consistency check is:
\begin{equation}
P(l|z,m) \propto \sum_{c\in C} \exp( -\alpha || \hat{z}_c(l)-z_c ||^2).
\end{equation}
Here, $l$ represents the estimated position, $z_c$ is the observation data and $\hat{z}_c$ is the prediction, $m$ stands for the map information, $C$ is the collection of cluster centers, and $\alpha$ is a tuning parameter.

(3) Distance validation and landmark matching: In addition to using pole-word consistency checks, we continue to employ traditional distance validation methods from Monte Carlo localization. By matching landmarks and calculating distances, we further verify and refine the position estimates. The combination of pole-words and traditional landmark matching significantly enhances the accuracy and robustness of the localization.

These advancements not only cater to the specific needs of navigating environments densely populated with pole-like structures but also contribute to the overall precision and reliability of the localization system.

\section{%
Experiments
}

\subsection{%
Dataset
}

We utilize the NCLT dataset to assess the proposed approach. The self-localization system's accuracy is enhanced by combining pole classification and self-localization. The University of Michigan campus provided two rounds of Segway robots for the acquisition of the NCLT (North Campus Long Term) dataset \cite{NCLT}. The NCLT dataset is well-suited for robotics research testing large-scale, long-term autonomous issues like long-term localization in urban environments. It can also be used for tasks like navigation and mapping in changing environments using LiDAR. The robot navigates the campus through iterative cycles, experiencing seasonal changes, diverse weather conditions (including rain, snow, and falling leaves), various times of day, indoor and outdoor environments, and encounters numerous static objects (such as trees, street signs, and light poles), as well as dynamic elements like cars, bicycles, and pedestrians. Additionally, it adapts to long-term structural changes caused by ongoing large-scale construction projects. The NCLT dataset's pole landmarks can be used as a good reference for this study's visual localization of poles. Even though the trajectories differ significantly between sessions, there is a significant amount of overlap. This suggests that in self-localization, a map-user robot can refer to the map to find valid landmarks from their overlapping areas.

\subsection{%
Localization Performance
}

This study utilizes an enhanced Monte Carlo particle filter approach for precise localization using environmental poles, integrating classification information of poles to optimize the filtering process. By correlating the class labels of poles with the state information of particles, our method significantly improves localization accuracy during particle weight updates by considering class matches.

In the experiments, the global map stores classified pole data corresponding to their class labels. The particle filter has been modified to incorporate pole words used to check consistency with the mapped poles' classes. During weight updates, if a class mismatch is detected, the affected particle's weight is multiplied by a small constant, thereby minimizing the impact of erroneous particles.

\newcommand{\val}[2]{${\mbox{#1}}_{\mbox{#2}}$}

\newcommand{\tabA}{
\begin{table*}
\begin{center}
\caption{Performance results.} \label{tab:A}
\begin{tabular}{|l|l|l|l|l|l|l|l|l|l|l|}
\hline
&
\multicolumn{2}{|c|}{\val{f}{map}(m)} &
\multicolumn{2}{|c|}{\val{$\Delta$}{pos}(m)} &
\multicolumn{2}{|c|}{\val{$\Delta$}{ang}(deg)} &
\multicolumn{2}{|c|}{\val{RMSE}{ang}(deg)} \\
\hline 
& ours & baseline
& ours & baseline
& ours & baseline
& ours & baseline
\\
\hline
2013-01-10 & 0.160 & 0.187 & 0.229 & 0.226 & 0.623 & 0.627 & 0.803 & 0.806 \\
2012-04-29 & 0.144 & 0.154 & 0.200 & 0.222 & 0.776 & 0.820 & 1.041 & 1.069 \\
2012-06-15 & 0.144 & 0.145 & 0.194 & 0.186 & 0.618 & 0.646 & 0.826 & 0.874 \\
\hline
\end{tabular}
\end{center}
\end{table*}
}

\tabA

\subsection{%
Implementation Details
}

Experiments were conducted using NVIDIA GeForce GTX 1080 GPUs, accommodating the computational needs of our deep learning models and particle filtering algorithms. For pole segmentation, we utilized the method detailed by Dong et al. \cite{dong2023online}. The feature extraction for pole-like landmarks was performed using the RIConv++ architecture as the back bone, a rotation-invariant neural network model described in Zhang et al. \cite{zhang2022riconv++} ensuring accurate classification despite orientation changes in 3D LiDAR data.

\subsection{%
Results
}

We evaluated the estimation errors of the baseline method and the proposed method.
Three independent sessions from the NCLT dataset, 
``2013-01-10,"
``2012-04-29,"
and ``2012-06-15",
are used to construct the training and test sets.
We used three independent sessions from the NCLT dataset, ``2013-01-10," ``2012-04-29," and ``2012-06-15," to construct training and test sets. Specifically, three experiments $i=$1, 2, and 3 were conducted. For each $i$-th experiment, the $i$-th dataset served as the test set, while the other two were used as training sets. Following the procedure outlined in paper \cite{dong2023online}, these two training sets were merged to construct a pre-built pole landmark map.

Table \ref{tab:A} shows performance results. We used mean and root mean square errors (RMSE) to the ground truth robot pose as performance metrics. We run the localization 10 times and compute the average means and RMSEs to the ground-truth trajectory. Our method is compared with the baseline in \cite{dong2023online}. Additionally, tests are conducted in various settings, including urban and suburban environments and across seasonal conditions, to evaluate the adaptability and robustness of the approach.

These results verify the fundamental effectiveness of the proposed method, which not only detects and segments landmarks but also classifies them.

\bibliography{ref} 
\bibliographystyle{IEEEtrans}

\end{document}